\documentclass[10pt,twocolumn,letterpaper]{article}

\usepackage{iccv}
\usepackage{times}
\usepackage{epsfig}
\usepackage{graphicx}
\usepackage{amsmath}
\usepackage{amssymb}

\usepackage{subfigure}
\usepackage{colortbl}
\usepackage{array}
\usepackage{xcolor}
\usepackage{multirow}
\usepackage{booktabs}
\usepackage{url}

\usepackage{comment}
\usepackage{color}
\newcommand{\cTab}[1]{{\color{red}{#1}}}

\newcommand{\cHiro}[1]{{\color{black}{#1}}}
\newcommand{\cNaka}[1]{{\color{black}{#1}}}
\newcommand{\cDel}[1]{{\color{black}{}}}
\newcommand{\cMatsu}[1]{{\color{black}{#1}}}
\newcommand{\cTBD}[1]{{\color{black}{#1}}}
\newcommand{\cYama}[1]{{\color{black}{#1}}}

\usepackage[breaklinks=true,bookmarks=false]{hyperref}

\iccvfinalcopy % *** Uncomment this line for the final submission

\def\iccvPaperID{****} % *** Enter the ICCV Paper ID here

\ificcvfinal\fi

\begin{document}

%%%%%%%%% TITLE
\title{Can Vision Transformers Learn without Natural Images?}

%\author{Kodai Nakashima^*, Hirokatsu Kataoka{*, Asato Matsumoto, Kenji Iwata\\
%National Institute of Advanced Industrial Science and Technology (AIST)\\
%Tsukuba, Ibaraki, Japan\\
%{\tt\small \{kodai.nakashima, hirokatsu.kataoka\}@aist.go.jp}
%\and
%Nakamasa Inoue\\
%Tokyo Institute of Technology\\

\author{
\renewcommand{\thefootnote}{\fnsymbol{footnote}}
Kodai Nakashima${}^\text{1}$\thanks {indicates equal contribution} \hspace{10mm} Hirokatsu Kataoka${}^\text{1}$$^*$\\
Asato Matsumoto${}^\text{1}$ \hspace{10mm} Kenji Iwata${}^\text{1}$  \hspace{10mm} Nakamasa Inoue${}^\text{2}$\\
National Institute of Advanced Industrial Science and Technology (AIST)${}^\text{1}$ \\ Tokyo Institute of Technology${}^\text{2}$\\
\tt\small {\{nakashima.kodai, hirokatsu.kataoka, matsumoto-a, kenji.iwata\}@aist.go.jp}
%\hspace{10mm} 
\\
\tt\small {inoue@c.titech.ac.jp}
}

\maketitle
\ificcvfinal\thispagestyle{empty}\fi

%%%%%%%%% ABSTRACT
\begin{abstract}
   Can we complete pre-training of Vision Transformers (ViT) without natural images and human-annotated labels? Although a pre-trained ViT seems to heavily rely on a large-scale dataset and human-annotated labels, recent large-scale datasets contain several problems in terms of privacy violations, inadequate fairness protection, and labor-intensive annotation. In the present paper, we pre-train ViT without any image collections and annotation labor. We experimentally verify that our proposed framework partially outperforms sophisticated Self-Supervised Learning (SSL) methods like SimCLRv2 and MoCov2 without using any natural images in the pre-training phase. Moreover, although the ViT pre-trained without natural images produces some different visualizations from ImageNet pre-trained ViT, it can interpret natural image datasets to a large extent. For example, the performance rates on the CIFAR-10 dataset are as follows: our proposal 97.6 vs. SimCLRv2 97.4 vs. ImageNet 98.0. The codes, datasets, and pre-trained models will be publicly available\footnote{\url{https://hirokatsukataoka16.github.io/Vision-Transformers-without-Natural-Images/}.}
\end{abstract}

%%%%%%%%% BODY TEXT
\section{Introduction}
% Ref (ViT): https://arxiv.org/abs/2010.11929
% Ref (DeiT): https://arxiv.org/abs/2012.12877

In contemporary visual recognition, a transformer architecture~\cite{VaswaniNIPS2017} is gradually replacing the \cHiro{usage} of convolutional neural networks (CNNs). The latter have been considered as central to the field of computer vision (CV). For example, the Residual Network (ResNet)~\cite{HeCVPR2016} is one of the de-facto-standard models in a wide range of visual tasks including image classification. The current high-standard scores on ImageNet are ResNet-based architectures, e.g., EfficientNet~\cite{TanICML2019,ForetICLR2021,PhamarXiv2020}, BiT~\cite{KolesnikovECCV2020}.

\begin{figure}[t]
    \centering
    \includegraphics[width=0.98\linewidth]{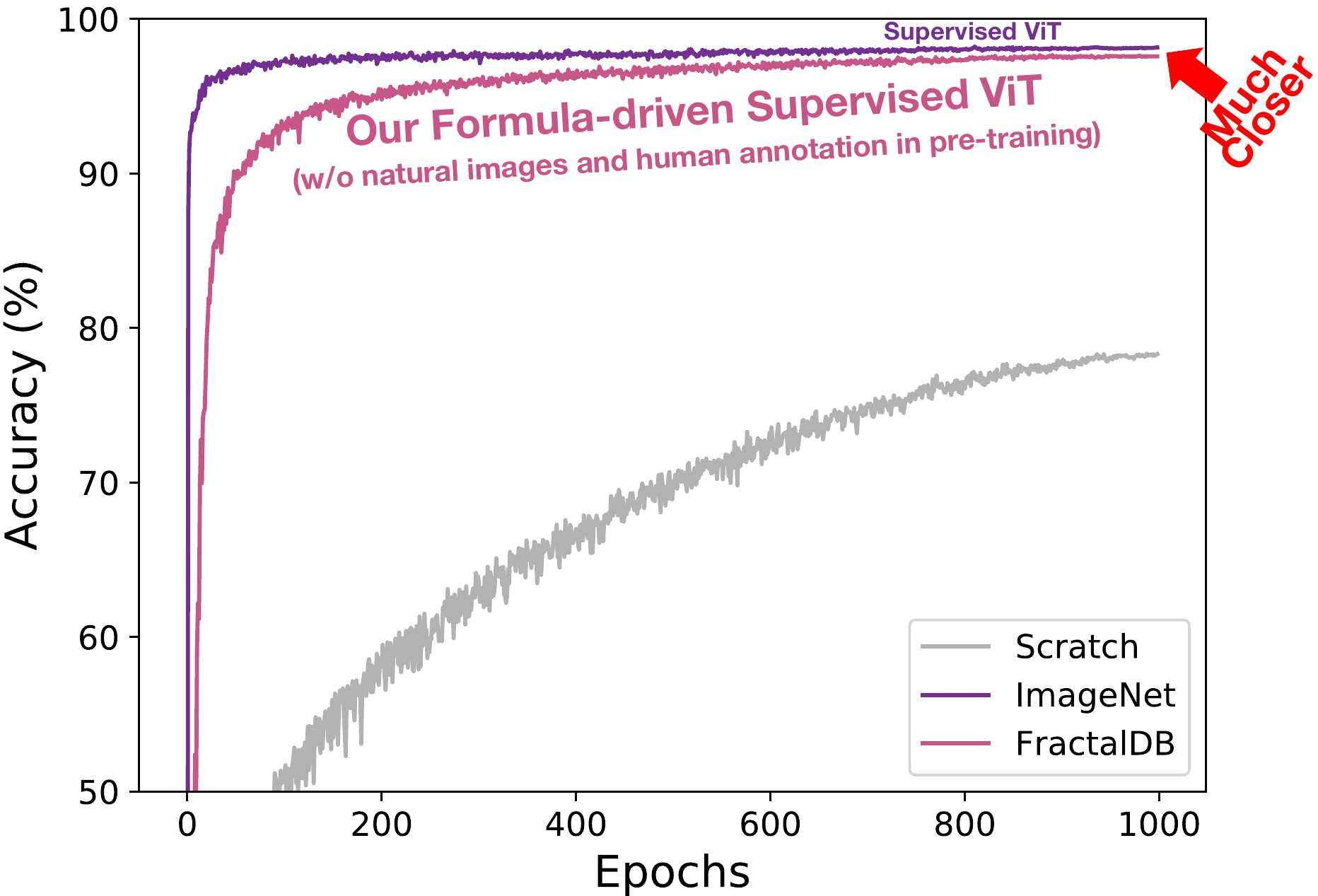}
    \caption{Accuracy transition in the fine-tuning phase. The graph illustrates that FractalDB-1k pre-trained ViT exhibits much higher training accuracy in early training epochs. The accuracy of FractalDB-1k is similar to that of ImageNet-1k pre-training.}
    \label{fig:accuracy_transition}
    \vspace{-10pt}
\end{figure}

\begin{figure*}[t]
    \centering
    \includegraphics[width=1.0\linewidth]{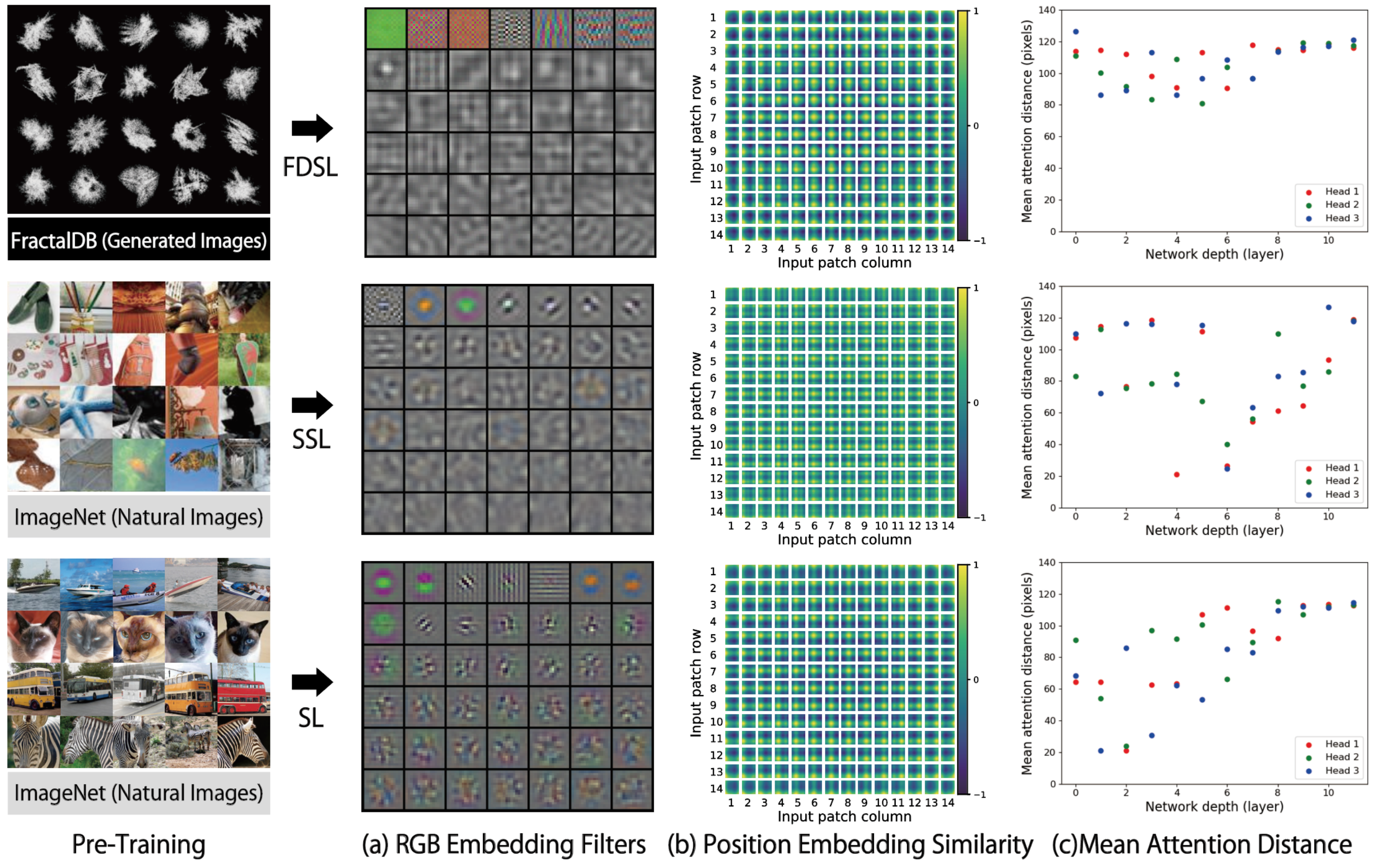}
    \vspace{-10pt}
%    \subfigure[Accuracy transition in the fine-tuning phase. The graph illustrates that FractalDB-1k pre-trained ViT exhibits much higher training accuracy in early training epochs. The accuracy of FractalDB-1k is similar to that of ImageNet-1k pre-training.]{\includegraphics[width=0.41\linewidth]{fig/c10_acc1.pdf} \label{fig:accuracy_transition}}
    \caption{\cNaka{Overview: We consider whether Vision Transformer (ViT) can pre-train without natural images.}
    %Our proposal: FractalDB pre-trained Vision Transformer (ViT). 
    \cHiro{Following the previous paper~\cite{DosovitskiyICLR2021}, we list the (a) RGB embedding filters, (b) position embedding similarity, and (c) mean attention distance in the frameworks of Formula-Driven Supervised Learning (FDSL) with FractalDB, Self-Supervised Learning (SSL) with SimCLRv2, and Supervised Learning (SL) with human-annotated ImageNet.} We replace pre-training based on human-labeled image datasets with \cMatsu{FDSL.} %←formula-driven supervised learning. 
    %Pre-training a \cNaka{model} with FractalDB. 
    %The framework with FDSL allows us to pre-train the ViT. Moreover, the FractalDB pre-trained ViT enables acquisition of unique feature representations while looking at the same position according to the similarity of positional embeddings.
    \cNaka{Compared to SSL and SL, FDSL pre-trained ViT enables the acquisition of slightly different filters, the same position embedding, and a wider receptive field.}
    }
    \label{fig:overview}
    \vspace{-10pt}
\end{figure*}

Transformer architecture, which consists of a self-attention mechanism, was initially employed in natural language processing (NLP) tasks such as machine translation and semantic analysis. We have witnessed the development of epoch-making methods, e.g., BERT~\cite{DevlinACL2019}, GPT-\{1, 2, 3\}~\cite{Radford2018,RadfordICML2018,BrownNeurIPS2020} with transformer modules. 
%A more recent study disclosed the rule as `Scaling Law' which is a simple knowledge in order to increase a performance rate in transformer~\cite{KaplanarXiv2020}.
The trend is gradually shifting from NLP to CV. One of the most active topics is undoubtedly Vision Transformers (ViTs) for image classification~\cite{DosovitskiyICLR2021}. ViTs effectively process and recognize an image based on transformers with minimum modifications. Even though the re-implementation is reasonably straightforward, it has been shown that ViTs often perform at least as well as state-of-the-art transfer learning. However, it is noteworthy that ViT architectures tend to require a large amount of data in the pre-training phase.
\cNaka{Dosovitskiy~\textit{et al.}~\cite{DosovitskiyICLR2021} reported that unless ViT is pre-trained with more than 100 million images, the accuracy is inferior to CNN.}
%However, ViT requires a huge amount of pre-training image data.
%In fact, ViT employed JFT-300M dataset~\cite{SunICCV2017} for constructing image pre-training model. 
The pre-training problem is somewhat alleviated by virtue of the Data-efficient image Transformer \cNaka{(DeiT)}~\cite{TouvronarXiv2020}.

%Here, we are not ready for infinite number of data increasing in transformer's Scaling Law. 
%Huge datasets such as JFT-300M and Instagram(IG)-3.5B~\cite{MahajanECCV2018} are not publicly available. Additionally, YFCC-100M was apparently withdrawn the right of dataset access\footnote{'Core Dataset' in \url{http://mmcommons.org/}}. 
On the other hand, using a large-scale image dataset may be problematic from the perspective of privacy preservation, annotation labor, and AI ethics. In fact, representative datasets including natural images taken by a camera are limited to academic \cHiro{or educational} usage. The problem cannot be solved even if we assign Self-Supervised Learning (SSL), e.g., MoCo~\cite{HeCVPR2020,ChenarXiv2020}, SimCLR~\cite{ChenICML2020,ChenNeurIPS2020}, and SwAV~\cite{CaronNeurIPS2020}, for automatic labeling of natural images. Training on natural image datasets still raises concerns in the context of privacy-violation and fairness-protection. Other large-scale datasets (e.g., Human-related images in ImageNet~\cite{YangFAT2020} and 80M Tiny Images~\cite{TorralbaTPAMI2008}\footnote{\url{https://groups.csail.mit.edu/vision/TinyImages/}}) are being deleted due to issues regarding AI ethics. To date, huge datasets such as JFT-300M and Instagram-3.5B~\cite{MahajanECCV2018} are not publicly available. Additionally, YFCC-100M has apparently withdrawn dataset access rights\footnote{`Core Dataset' in \url{http://mmcommons.org/}}. These dataset-related problems significantly limit the opportunities for research in this domain. The research community must carefully consider large-scale datasets in terms of availability and reliability while overcoming dataset-related problems.

%We believe that these aspects of large-scaleimage datasets and their pre-trained CNN models sig-nificantly  narrows down the  prospects  of vision-basedrecognition. The computer vision community must con-sider a large-scale dataset in terms of availability andreliability  while  overcoming  the  dataset-related  prob-lems.

%To overcome these dataset-related problems, 
In this context, Formula-Driven Supervised Learning (FDSL) was proposed in late 2020~\cite{KataokaACCV2020}. The concept involves automatically generating image patterns and their labels based on a mathematical `formula' which includes rendering functions and real-world rules. In the original paper, Kataoka~\textit{et al.} clarified that fractal geometry~\cite{Mandelbrot1983,fractals_everywhere} was the best way to construct a dataset in the framework of FDSL. Therefore, in the present paper, we consider whether ViTs \cYama{can be pre-trained} with FractalDB in FDSL. Although disadvantages of FractalDB pre-trained CNN have hitherto been pointed out\footnote{\url{https://www.technologyreview.com/2021/02/04/1017486/}}, we believe that the vision transformers can successfully \cYama{be pre-trained} with the FDSL framework because the self-attention mechanism enables elimination of the background effects between fractal and natural images, and can understand entire fractal shapes which consist of iteratively recursive patterns.
%Basically, the scale of dataset size in FDSL can be increased indefinitely, therefore, it might be a good match for transformer architecture under the theory of Scaling Law. 
Figure~\ref{fig:accuracy_transition}, \ref{fig:overview} illustrate the accuracy transition and characteristics of training properties. 
% As a matter of fact, the FractalDB pre-trained ViT performs similarly to human-annotated ImageNet pre-trained ViT \cHiro{(see also Table~\ref{tab:comparison})}, and slightly better than self-supervised ImageNet with SimCLRv2 pre-trained ViT \cHiro{(see also `Average' in Table~\ref{tab:fdsl_vs_ssl})}. %from the visualization.
%while acquiring unique features from the visualization.
% Therefore, we would like to investigate a larger FractalDB, more than 10k categories, in the experimental section. 

The contributions of the paper are as follows: We clarify that the FractalDB under the \cHiro{FDSL framework}  is more effective for ViT compared to CNN. The performance of FractalDB-10k pre-trained ViT is similar to those approaches with supervised learning \cHiro{(see Table~\ref{tab:comparison})}, and slightly surpasses the self-supervised ImageNet with SimCLRv2 pre-trained ViT \cHiro{(see `Average' in Table~\ref{tab:fdsl_vs_ssl})}. \cHiro{Here,} on the CIFAR-10 dataset the scores are as follows: FractalDB 97.6 vs. SimCLRv2 97.4 vs. ImageNet 98.0. Importantly, the FractalDB pre-trained ViT does not require any natural images \cHiro{and} human annotation in the pre-training.
%%%%%%%%%%%%%%%%%%%%%%%%%%%%%%%%%%%%%%%%%%%%%%%%%%%%%%%%%%%%%

% \clearpage

\section{Related work}

\cHiro{We would} like to discuss a couple of topics in visual transformers and pre-training datasets. \cHiro{We} mainly focus on architectures and large-scale datasets for image classification.

\subsection{Network Architectures for Image Recognition}

Convolutional Neural Networks (CNN) are popular in visual recognition. Several well-defined structures have emerged through a large number of trials in this decade~\cite{KrizhevskyNIPS2012,SimonyanICLR2015,SzegedyCVPR2015,HeCVPR2016,XieCVPR2017,HuangCVPR2017,TanICML2019}. Very recently, at the end of 2020, the architecture shifted to transformers~\cite{VaswaniNIPS2017} originating from natural language processing. Transformers basically consist of several modules with multi-head self-attention layers and \cMatsu{Multi-Layer  Perceptron} blocks. The mechanism has enabled the construction of revolutionary models (e.g., BERT~\cite{DevlinACL2019}, GPT-\{1, 2, 3\}~\cite{Radford2018,RadfordICML2018,BrownNeurIPS2020}). Thus, the computer vision community is focusing on replacing the de-facto-standard convolutions with a transformer-based architecture. One of the most insightful architectures is the Vision Transformer (ViT)~\cite{DosovitskiyICLR2021}. Though the ViT is a basic transformer architecture in terms of image input, the model performs comparably to state-of-the-art alternative approaches on several datasets. However, ViT requires over ten-million-order labeled images in representation learning. The JFT-300M/ImageNet-21k pre-trained ViT was verified by experiments to perform well in terms of accuracy. The issue of learning with large-scale datasets was alleviated with the introduction of the Data-efficient image Transformer (DeiT)~\cite{KaplanarXiv2020}. However, the pre-training problem still remains in image classification.

%CNNs: AlexNet~\cite{KrizhevskyNIPS2012}, GoogLeNet Inception~\cite{SzegedyCVPR2015}, VGG~\cite{SimonyanICLR2015}, ResNet~\cite{HeCVPR2016}, ResNeXt~\cite{XieCVPR2017}, DenseNet~\cite{HuangCVPR2017}, EfficientNet~\cite{TanICML2019},

%Transformers: ViT~\cite{DosovitskiyICLR2021}, DeiT~\cite{KaplanarXiv2020}, T2T-ViT~\cite{yuanarXiv2020}

%Others (NLP): Original Transformer~\cite{VaswaniNIPS2017}, BERT~\cite{DevlinACL2019}, GPT-\{1, 2, 3\}~\cite{Radford2018,RadfordICML2018,BrownNeurIPS2020}

\subsection{Image dataset and training framework}

It is said that the deep learning era started from ILSVRC~\cite{RussakovskyIJCV2015}. Undoubtedly, transfer learning with large-scale image datasets has contributed to accelerating  visual training~\cite{HeICCV2019}. Initially, the ImageNet~\cite{DengCVPR2009} and Places~\cite{ZhouTPAMI2017} pre-trained models were widely used for diverse tasks, not limited to image classification. \cYama{However}, even in million-scale datasets, there exist several concerns such as AI ethics and copyright problems, e.g., fairness protection, privacy violations, and offensive labels. Due to these sensitive issues, as mentioned above, human-related labels in ImageNet~\cite{YangFAT2020} and 80M Tiny Images~\cite{TorralbaTPAMI2008} were deleted. We must pay attention to the terms of use in large-scale image datasets and create pre-trained models accordingly.

On one hand, to alleviate the image labeling labor required of human annotators, Self-Supervised Learning \cMatsu{(SSL)} progressed significantly in recent years. The early methods created pseudo labels based on semantic concepts~\cite{DoerschICCV2015,NorooziECCV2016,NorooziCVPR2018,NorooziICCV2017,GidarisICLR2018} and trained feature representations through image reconstruction~\cite{ZhangECCV2016}. By contrast, the SSL methods are closer to supervised learning with human annotations in terms of performance rates (e.g., MoCo~\cite{HeCVPR2020,ChenarXiv2020}, SimCLR~\cite{ChenICML2020,ChenNeurIPS2020}, SwAV~\cite{CaronNeurIPS2020}).
In this context, Formula-Driven Supervised Learning (FDSL)~\cite{KataokaACCV2020} was proposed to overcome the problems of AI ethics and copyrights in addition to annotation labor. The framework is similar to self-supervised learning. However, FDSL methods do not require any natural images taken by a camera. The framework simultaneously and automatically generates image patterns and the paired labels for pre-training image representations. We would like to investigate whether the formula-driven image dataset can sufficiently optimize a vision transformer in the pre-training phase. At the same time, we will compare the pre-trained ViT through the FDSL framework with supervised and self-supervised pre-training. If the supervised/self-supervised pre-training can be replaced by FDSL, vision transformers may be pre-trained without using any natural images in the future.

\section{Vision transformer (ViT)}

As mentioned in the previous sections, we believe that FractalDB pre-training can replace pre-training with natural image datasets by combining with ViT architecture. The FDSL framework including FractalDB enables automatic generation of an infinite number of training categories and their image labels with a mathematical formula. In the ViT characteristics, the FractalDB pre-trained ViT must be better than CNN. Additionally, although FractalDB does not contain a background area inside of the image, the self-attention mechanism effectively focuses on the fractal patterns while ignoring background areas.
Moreover, the FractalDB pre-trained ViT is also better than the pre-training with natural image datasets in terms of privacy protection, AI ethics, and annotation labor. Here, we explore the potential of the transformer in visual tasks.

%In model design we follow the original Transformer (Vaswani et al., 2017) as closely as possible.An advantage of this intentionally simple setup is that scalable NLP Transformer architectures – andtheir efficient implementations – can be used almost out of the box

The basic transformer requires a 1D sequence of tokens in the input layer. To process 2D images, the image $x \in \mathbb{R}^{H \times W \times C}$ is reshaped into flattened image patches $x_p \in \mathbb{R}^{N \times (P^2 \cdot C)}$, where $\left(H, W\right)$ is the original image size, $C$ is the number of channels, $\left(P, P\right)$ is the size of each image patch, and $N = HW/P^2$ is the number of patches. \cNaka{Flattened image patches are converted into D-dimensional vectors by a trainable linear projection and processed in a fixed dimension through all layers.} After the linear projection, adding trainable 1D position embeddings to the patch representation \cNaka{and concatenate \cNaka{classification token} similar to BERT} becomes the input sequence of the transformer encoder.

The transformer encoder block consists of the multi-head self-attention layer and the \cNaka{Multi-Layer Perceptron (MLP)}. The self-attention, called scaled dot-product attention, firstly computes a set of queries $Q = XW_{Q}$, a set of keys $K = XW_{K}$, and a set of values $V = XW_{V}$, in which $X \in\mathbb{R}^{(N+1) \times D}$ is an input sequence, $W_{Q} \in \mathbb{R}^{D \times d}$, $W_{K} \in \mathbb{R}^{D \times d}$, $W_{V} \in \mathbb{R}^{D \times d}$ are trainable weights, and $d$ is vector size. \cHiro{The self-attention is computed as follows:} %The output of the self-attention is computed as follows:
\begin{align}
    {\rm Attention}(Q, K, V) = {\rm softmax}\left(\displaystyle\frac{QK^{\mathrm T}}{\sqrt{d}}\right)V
\end{align}
where the softmax function is applied over each row of the matrix. In multi-head self-attention (MSA), $h$ heads are added to self-attention as follows:
\cNaka{
\begin{align}
    {\rm MSA}(Q, K, V) &= {\rm concat}(head_{1}, ..., head_{h})W \\
    head_{h} &= {\rm Attention}(Q_{h}, K_{h}, V_{h})
\end{align}
}
Each head provides a sequence of size \cNaka{$(N+1) \times d$}. These $h$ sequences are rearranged into an \cNaka{$(N+1) \times dh$} sequence that is reprojected by an MLP into \cNaka{$(N+1) \times D$.} In summary, the transformer encoder processes the image as follows:
\cNaka{
\begin{align}
    % z_0 &= \left[x_{\rm{class}}; x_p^1{\rm{E}}; x_p^2{\rm{E}}; ... ; x_p^N{\rm{E}}\right] + \rm{E}_{pos} \\
    z_0 &= \left[x_{\rm{class}}; {\rm MLP}(x_p^1); ... ; {\rm MLP}(x_p^N)\right] + {\rm E}_{pos} \\
    z'_l &= {\rm MSA}({\rm Norm}(z_{l-1})) + z_{l-1} \\
    z_l &= {\rm MLP}({\rm Norm}(z'_{l})) + z'_{l} \\
    y &= {\rm Norm}(z^0_L)
\end{align}
}
We conduct a characteristic evaluation on mechanisms in the ViT architecture such as linear embeddings, positional embeddings, and attention maps. Though the detailed characteristics are described and visualized in the experimental section, we generated interesting results. For example, the filters of the first linear embedding in FractalDB pre-training are different from the ImageNet pre-trained model \cNaka{(see Figure~\ref{fig:overview}\cTab{(a)})}. However, the positional embeddings \cNaka{(see Figure~\ref{fig:overview}\cTab{(b)})} are similar. Moreover, the FractalDB pre-trained ViT focuses on an object-specific area to understand the object in an image because of the rendering process without any background area \cNaka{(see Figure~\ref{fig:att})}. In the next section, we will explain data structure and generation for better understanding of FractalDB.

\section{Formula-Driven Supervised Learning}
%FDSL・FractalDB何かや関係性を説明したかった
%ここら辺でViTへの言及やACCV論文との差分はこれ、という部分が冒頭に必要
%ViTでFractalDBが有効な理由，accvとの差分をを説明したい
This section presents Formula-Driven Supervised Learning (FDSL) for vision transformers (ViT). We begin from a brief review of FractalDB~\cite{KataokaACCV2020} under the framework of FDSL. We also describe how to apply the auto-generated pre-training dataset for ViT.

\subsection{Definition}
% The goal of Formula-Driven Supervised Learning (FDSL) is to accomplish pre-training without any natural images. In FDSL,  the framework automatically creates the paired image patterns and their labels by following a mathematical formula.
The goal of \cMatsu{FDSL} is to accomplish pre-training without any natural images. The framework automatically creates paired image patterns and their labels by following a mathematical formula.
Unlike the pre-training framework in supervised learning, the FDSL does not require any natural images \cHiro{and} human annotated labels.
More specifically, FDSL is formulated as follows: 
\begin{align}
    \label{eq:fdsl}
    \mathop{\rm argmax}_{M}\limits \mathbb{E}_{y,s} [ \ell(M(x), y) ]
    \; \mbox{s.t.}\; x = F(\theta,s),\; y = \theta ,
\end{align}
where $M$ is a network to be pre-trained, $\ell$ is a loss function, $x$ is a generated image pattern, and $y$ is a label in the image.
The image patterns are generated by a mathematical formula $F$, whose inputs are a parameter $\theta$ and a random seed $s$.
The network learns to predict the parameter $\theta$ used to generate $x$.
For simplicity, we assume that $y$ follows a uniform distribution over a pre-defined discrete set of parameters $\Theta = \{\theta_{k}\}_{k}^{K}$.
This allows us to introduce $K$-class classification loss such as cross-entropy loss for $\ell$.

%\cTBD{A strength of FDSL is that it can \cHiro{theoretically} generate an infinite number of image patterns with labels. Therefore, it would be effective for vision transformers which require a huge number of images for pre-training.}

\subsection{FractalDB}
% FractalDBのcategoryとinstanceの作り方について説明
% 特にinstanceの増やし方について詳細・具体的に言及
% accv 3.Automatically generated large-scale datasetを参考に作成 
One of the most successful approaches in FDSL relies on fractals.
FractalDB consists of 1k to 10k pairs of fractal images generated with the iterated function system (IFS)~\cite{fractals_everywhere}.
The reason that fractal geometry is chosen to generate the dataset is that the function can render complex patterns \cYama{and different shapes for each parameter set.}
%that exist in natural phenomena and have a different shape for each parameter set.

%and probability of occurrence  The category of the image is given by the parameters.
In Equation~\ref{eq:fdsl}, $F$ and $\theta_i$ correspond to IFS and $(a_i, b_i, c_i, d_i, e_i, f_i, p_i)$, respectively. The parameters are randomly searched and will be adopted when the image patterns generated from the parameters exceed the threshold of the filling rate which is calculated by dividing the number of pixels of the fractal dot by the total number of pixels of the image. The intra-category instances are expansively generated by three methods for considering category configurations to maintain the shape in the category: varying the parameters slightly, rotation, and drawing with patch. Varying the parameters is the process of multiplying one of the 6 parameters of IFS by weights. We can generate the image from this parameter, which changes the detailed representation while maintaining the general shape of the category. 
By multiplying one of each parameter by 4 weights, 25 (original 1 + params 6 $\times$ weights 4) different variations of the image were generated. In the second method, rotation, we manipulate the flipping operation in the image. There are 4 rotations \{none, horizontal flip, vertical flip, horizontal vertical flip\}. Drawing with patch is the process of rendering the fractal image with patch instead of point. In FractalDB, 10 different 3 $\times$ 3 [pixel] patches were used to generate the fractal image. Finally, adopting all three methods can create 1,000 (25 $\times$ 4 $\times$ 10) intra-category instances.

The basic FractalDB consists of 1,000 or 10,000 different fractal categories and 1,000 instances. In experiments, the ResNet-50 as CNN model pre-trained with FractalDB partially outperformed models pre-trained with human-annotated datasets such as ImageNet and Places. 

\subsection{FractalDB for Vision Transformers}
% ACCVとの差分
% In the present paper, we introduce two improvements to FractalDB: (i) Larger categories, and (ii) Colored fractal images. They help to enhance pre-training vision transformers because xxx
We introduce two \cHiro{modifications} to FractalDB \cHiro{pre-trained models according to the architecture specifications}: \cHiro{(i) Colored fractal images and (ii) Training epoch. We investigate whether the performance rate \cYama{of FractalDB pre-trained ViT improves or not} with these configurations by following the success of self-supervised learning with natural images. We believe that it is necessary to utilize colored images for pre-trainig to recognize natural images in a longer training time.} 

%(i) Larger categories, and (ii) Colored fractal images. 
%These improvements help to enhance pre-training vision transformers because \cTBD{xxx}
%we improved the FractalDB for more effective learning of ViT. In this way, we aim to obtain a more generic feature representation of the model.
    
%\noindent {{\bf Larger categories.}}
% カテゴリ数をさらに増強したFractalDB-30k/50k/100kを作成した．これにより10kよりも画像枚数が3/5/10倍になり，Scaling LawからViTの学習に有効だと考えられる．
% カテゴリ数を増やすときの工夫点等を加えたい
%The previous dataset has up to 10k categories. We generated FractalDB-30k/50k/100k of which the number of categories was extended by the feature of Formula-driven image datasets to generate an infinite number of images. It is expected that the model pre-trained with the FractalDB which has a larger number of images will acquire valid feature representations because of the Scaling Law.

\noindent {\bf Colored fractal image.}
% 色付きのフラクタル画像を生成した．自然画像は色がついていることが一般的である．これまでのFractalDBはグレイスケール画像であり，そこに差があった．フラクタル画像を生成する際に点・パッチを描画する毎に，その点・パッチのランダムで色付けを行った．これを学習することで色も考慮した特徴表現の獲得も可能となる．
Images of conventional FractalDB were drawn by moving dots or patches in grayscale. However, the natural images for common pre-training are not only grayscale, but also various color combinations. The model pre-trained with the datasets constructed natural images has representations related to color distribution in nature~\cite{zeiler2014visualizing}. Therefore, we generated the FractalDB in color. The generating procedure was to draw points or patches colored randomly each \cHiro{iteration} time. By pre-training with a dataset of colored fractal images, the model acquires feature representations related to color.

\noindent \cHiro{{\bf Training epoch.} The recent \cMatsu{Self-Supervised Learning (}SSL) methods consider a longer training. For example, SimCLR tried a longer training epochs up to 1k [epoch]~\cite{ChenICML2020}. Therefore, although the first work in FDSL~\cite{KataokaACCV2020} conducted with only 90 [epoch], we further verify a suitable training term. Therefore, we also plan to implement a longer training epoch with reference to the recent SSL methods. Here, in the experimental section, we evaluate up to 300 epochs for a further improvement in the FractalDB pre-trained ViT.}

\subsection{Explore parameters}
\label{sec:explore_parameters}
\begin{figure*}[t]
    \centering
    \subfigure[\cMatsu{C10}]{\includegraphics[width=0.24\linewidth]{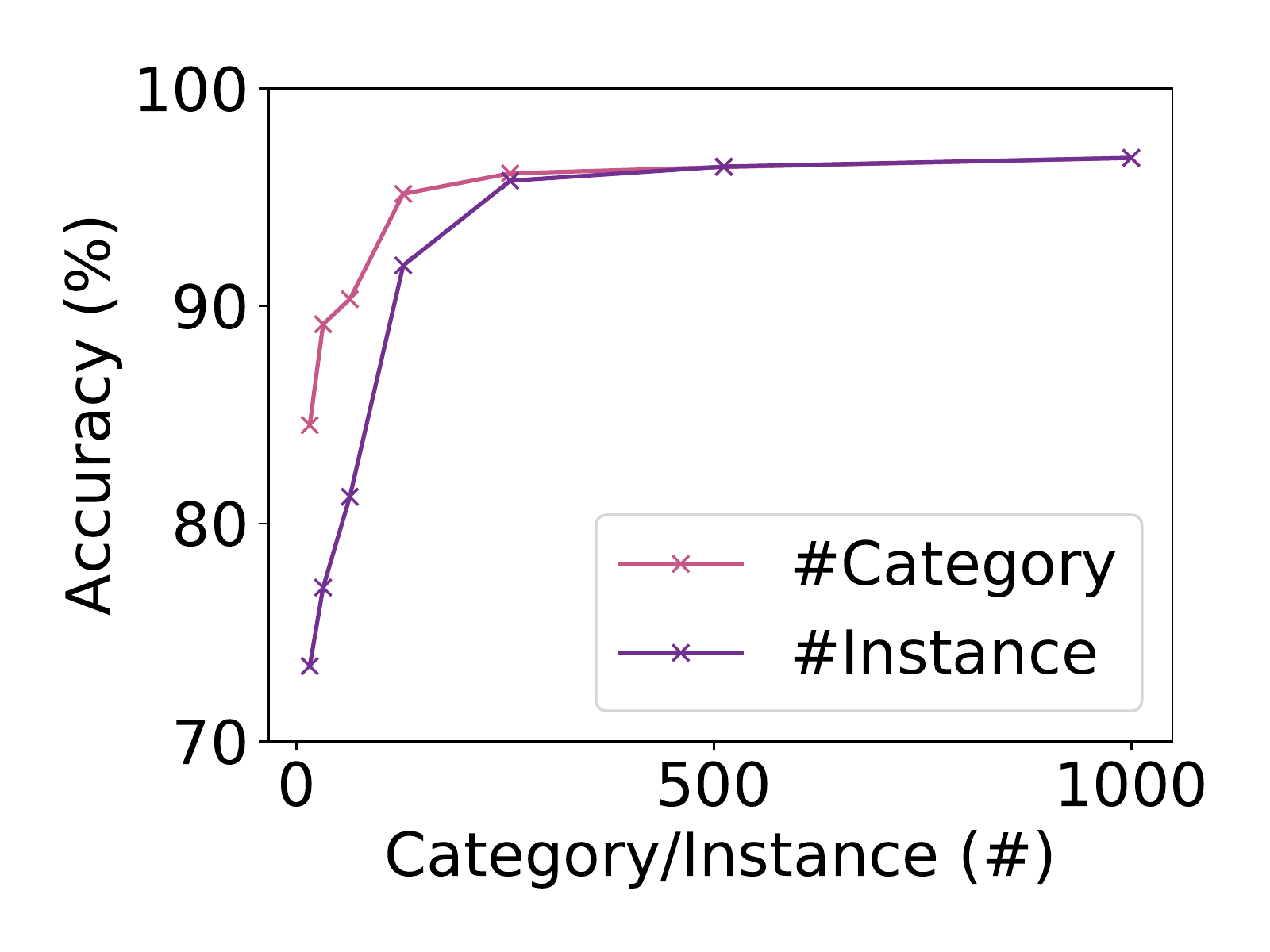} \label{fig:catins_c10}}
    \subfigure[\cMatsu{C100}]{\includegraphics[width=0.24\linewidth]{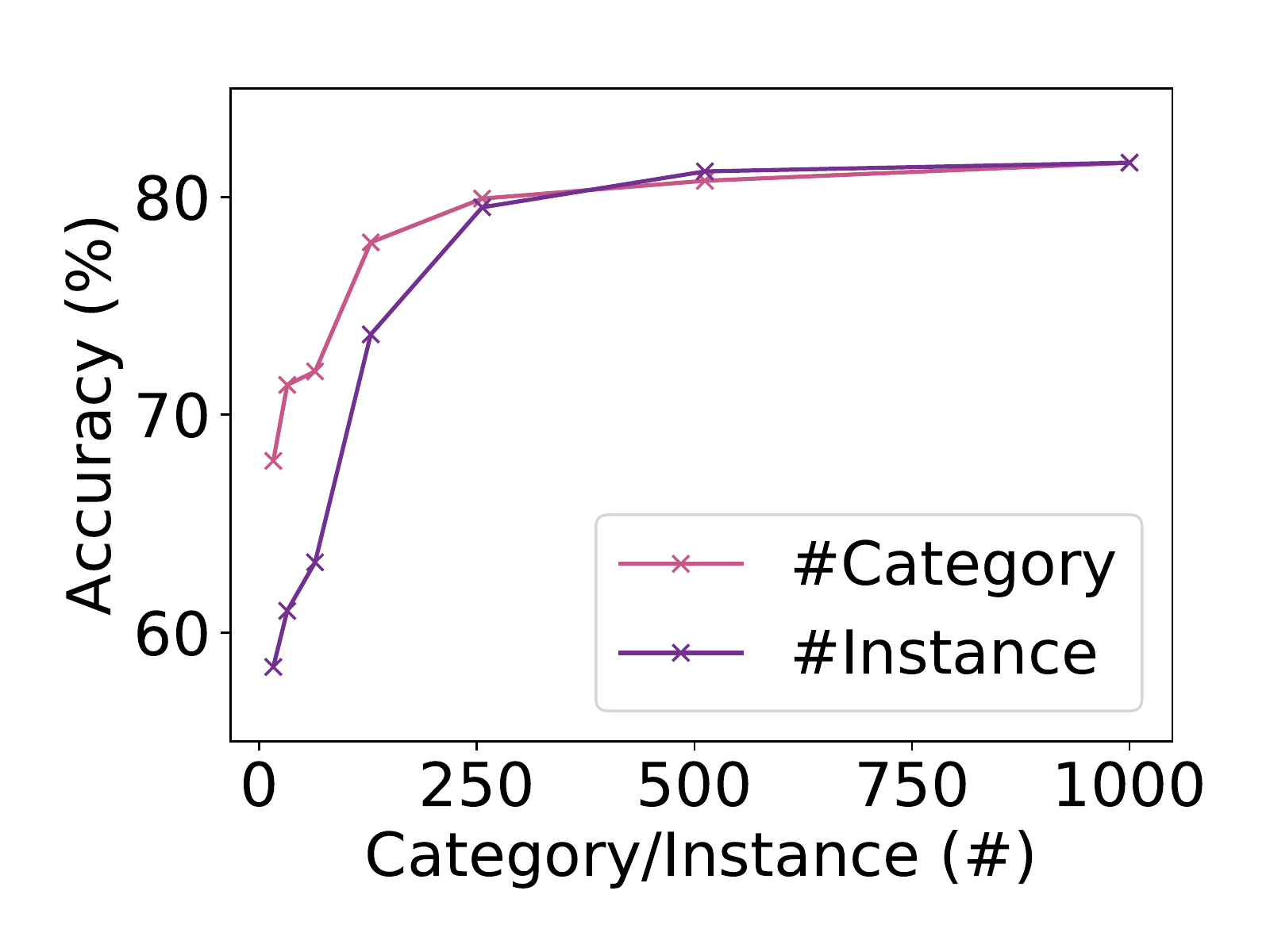} \label{fig:catins_c100}}
    \subfigure[Cars]{\includegraphics[width=0.24\linewidth]{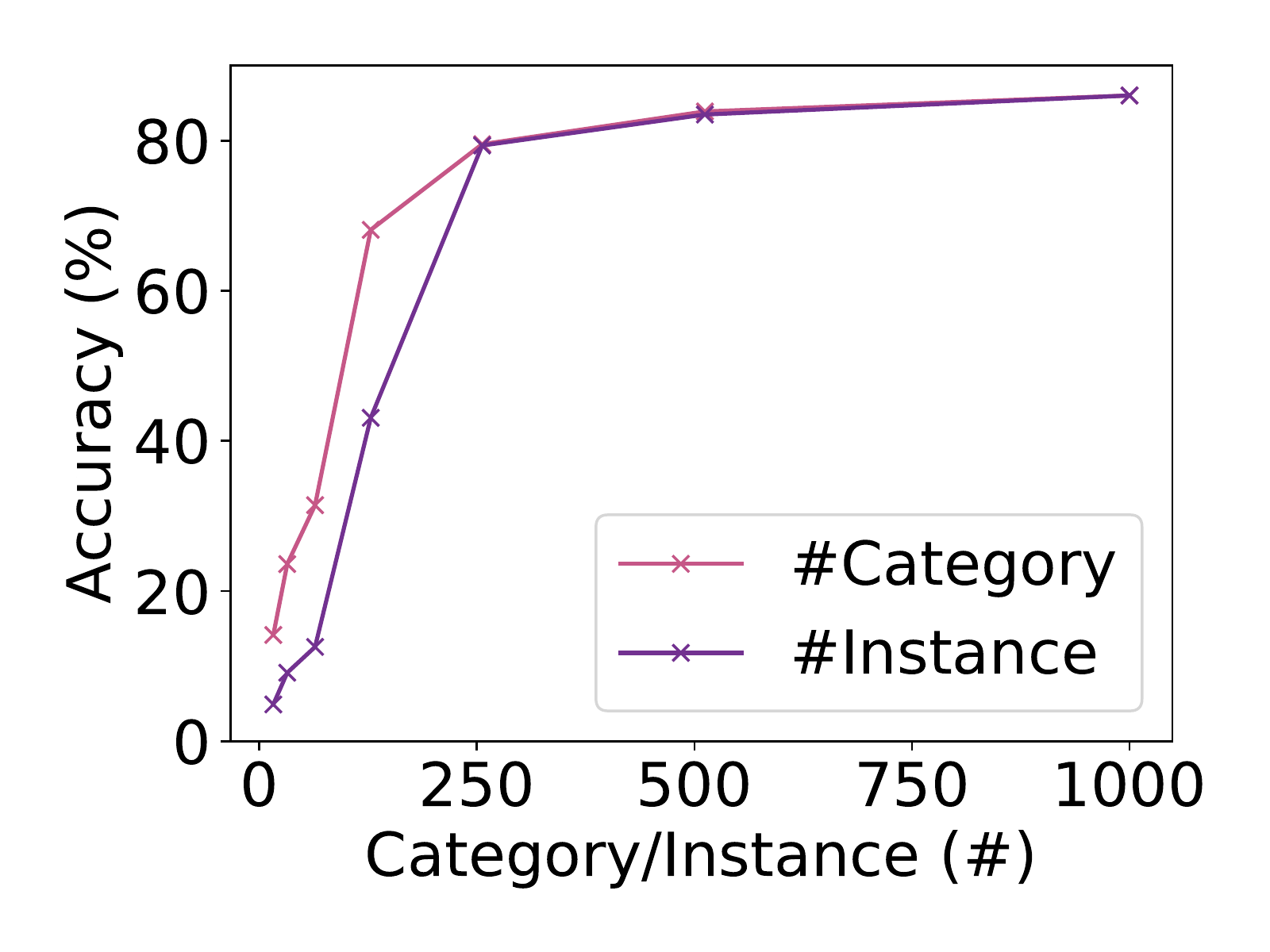} \label{fig:catins_cars}}
    \subfigure[Flowers]{\includegraphics[width=0.24\linewidth]{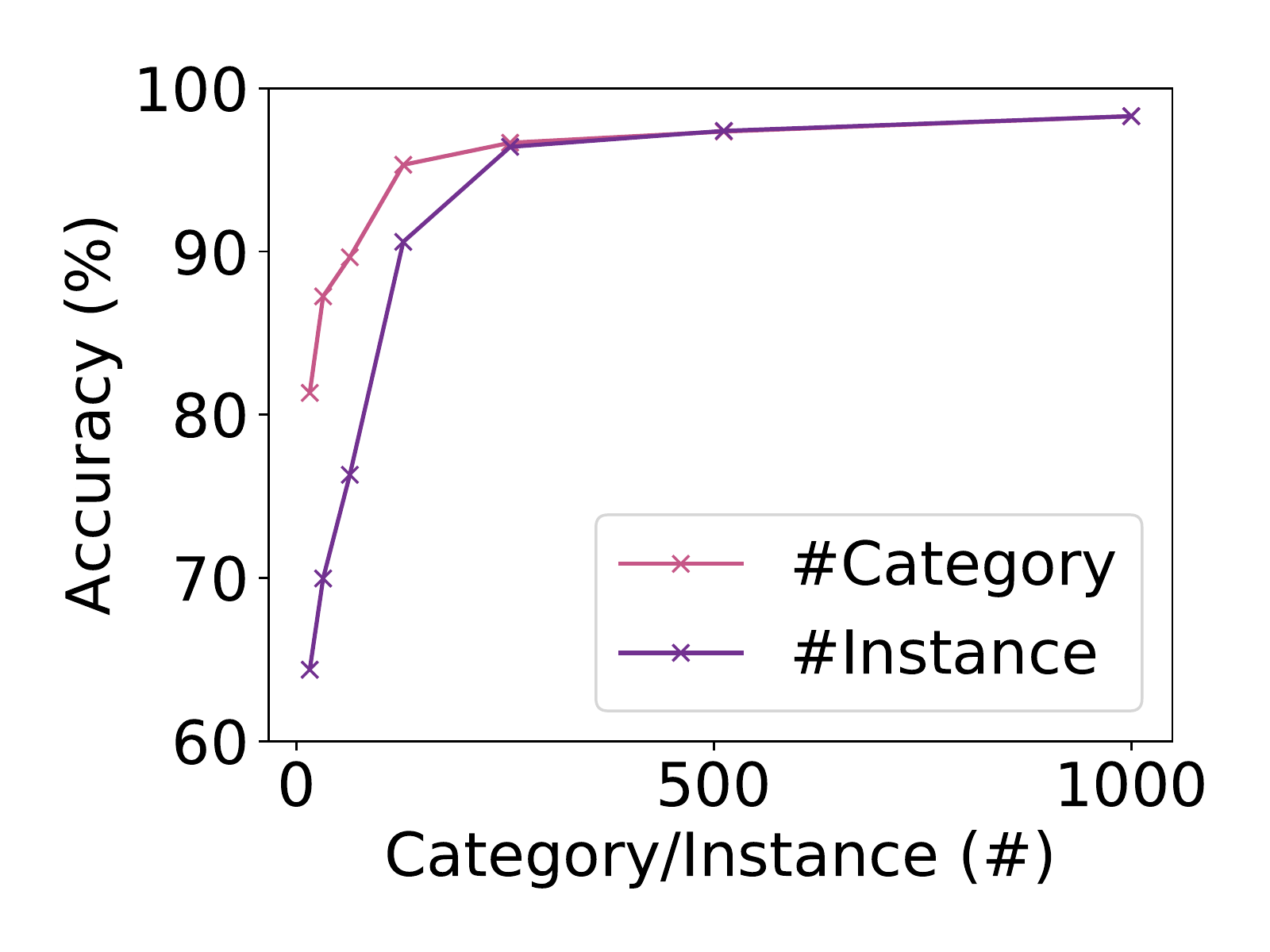} \label{fig:catins_flowers}}
    \caption{\cMatsu{Effects of \#category and \#instance.} The other parameter is fixed at 1,000, e.g., \#Category is fixed at 1,000 as \#Instance varies among \{16, 32, 64, 128, 256, 512, 1,000\}.}
    \label{fig:catins}
    \vspace{-13pt}
\end{figure*}

% accvで行ったパラメータ探索実験についての紹介，5.1節につなげたい
In FractalDB, the parameters related to the configuration of the dataset and image generation methods were experimentally investigated. Kataoka \textit{et al.}~\cite{KataokaACCV2020} explored \#category and \#instance, patch vs. point, filling rate, weight of intra-category fractals, \#dot, and image size. 
%The accuracy of fine-tuning is affected by the composition of the dataset used for pre-training. 
\cHiro{Here, we only investigate effective parameters for exploration study in ViT architecture. According to their study, we further carry out the experiments in terms of \#category/\#instance (see Figure~\ref{fig:catins}), 1k/10k categories (see Table~\ref{tab:larger_category}), patch vs. point (see Table~\ref{tab:patchvspoint}), in addition to above-mentioned grayscale vs. color (see Table~\ref{tab:grayscale_vs_color}) and training epochs (see Table~\ref{tab:training_epochs}). At the same time, we first compare FractalDB pre-training with other FDSL frameworks and training from scratch (see Table~\ref{tab:comparison_fddb}) and evaluate patch size which is one of the important parameters in ViT (see Table~\ref{tab:patchsize}).}

%They studied the effects of changing the number of categories and instances. Not only the configuration of the dataset but also the appearance of images used for pre-training is considered to be important for acquiring the model representing a generic feature. Therefore, Kataoka \textit{et al.} also conducted studies changing hyper-parameters related to the representation of the image such as patch vs. point and weights, among other things. In exploring different weights of intra-category fractals, the intervals of weights were varied as follows: \{0.8, 0.9, 1.0, 1.1, 1.2\} or \{0.01, 0.5, 1.0, 1.5, 2.0\}.

%They confirmed that the model pre-trained with FractalDB containing large numbers of categories and instances tends to perform well in terms of accuracy in fine-tuning. The following conditions were found to be superior for generating fractal patterns: drawing is the patch with 3 $\times$ 3, the filling rate is 0.10, the weight is 0.4, \#dot is 200k, and the image size is 256 $\times$ 256 or 362 $\times$ 362 [pixel]. In the present paper, we investigated the parameters of \#category and \#instance, patch vs. point, and patch size in ViT in Section~\ref{subsec:exploration}.

% DeiT-T/16 #cat/#ins

\section{Experiments}
% 実験では，はじめにViTに有効なFractalDBの構成方法を検討する．次に，FractalDBを自然画像データセット（ImageNet-1K，Places365）と，ファインチューニング時の精度で定量的に評価し比較する．
We verify the effectiveness of FractalDB pre-trained ViT in multiple respects. First, we explore a better configuration of FractalDB for ViT. Then we evaluate the best configuration in FractalDB pre-trained ViT on several image datasets, namely CIFAR-10/100 (C10/C100), Stanford Cars (Cars), and Flowers-102 (Flowers), by following the paper~\cite{KaplanarXiv2020}. Moreover, we quantitatively compare the FractalDB pre-trained ViT with the pre-training with representative large-scale image datasets (e.g., ImageNet-1k, Places-365) and architectures (e.g., ResNet-50).
%In a set of experiments, we first investigate how to construct an effective FractalDB for ViT. We then quantitatively evaluate and compare the FractalDB with natural image datasets (ImageNet-1k and Places365).

% FractalDBの特性を確認し，事前学習で獲得した特徴表現を先行研究と比較するためにViTモデルを使用した．我々は，基本的に先行研究でImageNetで調整された学習設定に従うが，ViTは学習率に敏感であるため{0.0001, 0.0003, 0.0005}の中から学習が完了した学習率の最大値を採用した．
Here, to confirm the properties of the FractalDB pre-trained \cHiro{model},
%and compare the pre-trained feature representation with previous work,
we simply use the original ViT model (more specifically, we assign DeiT\cHiro{; hereafter, we assign DeiT for the experiments in the paper}) without any modification. We investigate the pre-training method with various parameters. \cTBD{For example, we explore different learning rates
%, specifically among \{0.0001, 0.0003, 0.0005\}, 
in the pre-training phase since \cHiro{DeiT} is known to be \cYama{an architecture parameter-sensitive to different training datasets}.}
The fine-tuning setting is the same as that of ~\cite{KaplanarXiv2020}.
%On the other hand, in the fine-tuning phase, we adjust all parameters on all datasets by following the previous study~\cite{KaplanarXiv2020}.

%
%We basically pre-train and fine-tune according to the learning settings adjusted by ImageNet in previous studies~\cite{KaplanarXiv2020}. 
%ViT is sensitive to the learning rate, and depending on the dataset, pre-training may not proceed. 
%Therefore, we pre-trained with learning rates of \{0.0001, 0.0003, 0.0005\}  for each dataset, and adopted the largest value among those that could be learned.

% \begin{table}[t]
%     \begin{center}
%     \caption{write caption here...}
%     \begin{tabular}{cc|cccc} \hline
%          Patch & Color & C10 & C100 & Cars & Flowers \\\hline
%          &  & 94.2 & 77.3 & 65.4 & 95.1 \\
%          \checkmark & & \cTBD{**.*} & \cTBD{**.*} & 87.1 & 98.3 \\
%          \checkmark & \checkmark & 96.8 & 81.6 & 86.0 & 98.3 \\\hline
%     \end{tabular}
%     \end{center}
% \end{table}

% DeiT-T/16でのFDDB事前学習モデル比較実験
\begin{table}[t]
    \begin{center}
    \caption{Comparisons of ViT pre-training among FractalDB and other formula-driven image datasets with Bezier curves (BezierCurveDB) and Perlin noise (PerlinNoiseDB).}
    \begin{tabular}{lcccc} \toprule[0.8pt]
         & C10 & C100 & Cars & Flowers \\ \midrule[0.5pt]
        Scratch & 78.3 & 57.7 & 11.6 & 77.1 \\
        \cNaka{PerlinNoiseDB} & 94.5 & 77.8 & 62.3 & 96.1 \\
        \cNaka{BezierCurveDB} & 96.7 & 80.3 & 82.8 & \textbf{98.5} \\
        \rowcolor[gray]{0.8}%
        FractalDB-1k & \textbf{96.8} & \textbf{81.6} & \textbf{86.0} & 98.3 \\ \bottomrule[0.8pt]
    \end{tabular}
    \label{tab:comparison_fddb}
    \end{center}
    \vspace{-13pt}
\end{table}

% DeiT-T/16での自然画像学習モデルとの比較実験．現状のスコアを掲載．
\begin{table}[t]
    \begin{center}
    \caption{Exploration of larger categories on FractalDB. We compare FractalDB-1k pre-training with FractalDB-10k pre-training.}
    \begin{tabular}{lcccc} \toprule[0.8pt]
        Pre-train \#cat & C10 & C100 & Cars & Flowers \\ \midrule[0.5pt]
        %Scratch & -- & 78.3 & 57.7 & 11.6 & 77.1 \\
        %Places-365 & Natural & 97.6 & 83.9 & 89.2 & 99.3 \\
        %ImageNet-1k & Natural & 98.0 & 85.5 & 89.9 & 99.4 \\ \hline
        1k & 96.8 & 81.7 & 86.0 & 98.3 \\
        10k & \textbf{97.6} & \textbf{83.5} & \textbf{87.7} & \textbf{98.8} \\ \bottomrule[0.8pt]
        %30k (500ins) & -- & -- & -- & -- \\ 
        %50k (500ins) & -- & -- & -- & -- \\ 
        %100k (500ins) & -- & -- & -- & -- \\ \hline
    \end{tabular}
    \label{tab:larger_category}
    \end{center}
    \vspace{-20pt}
\end{table}

\subsection{Exploration study} \label{subsec:exploration}
% このサブセクションでは，我々は[]を参考にViTに効果的なFractalDBの構成を検討する．
We explore an effective FractalDB configuration for \cHiro{DeiT} under the reference in Kataoka \textit{et al.}~\cite{KataokaACCV2020}. According to the previous work, a full exploration would be very time-consuming. Therefore, we seek to implement the most influential parameters \cHiro{which are described in Section~\ref{sec:explore_parameters}.} %, namely comparisons with other formula-driven image datasets (see Table~\ref{tab:comparison_fddb}), \#category, \#instance (see Figure~\ref{fig:catins}), larger category size (see Table~\ref{tab:larger_category}), and patch vs. point (see Table~\ref{tab:patchvspoint}). Moreover, we carry out the experiments on colorized FractalDB (see Table~\ref{tab:grayscale_vs_color}) and ViT-specified exploration study with patch size (see Table~\ref{tab:patchsize}).

\noindent {{\bf Comparison with other formula-driven image datasets (see Table~\ref{tab:comparison_fddb}).}}
In addition to FractalDB, Kataoka et al.~\cite{KataokaACCV2020} also proposed datasets based on Perlin noise (PerlinNoiseDB) and Bezier curves (BezierCurveDB). We try to pre-train and fine-tune with these formula-driven image datasets on the \cHiro{DeiT} architecture. This allows us to determine whether the proposed FractalDB performs the best through pre-training.  
From Table~\ref{tab:comparison_fddb}, we can confirm the existence of higher accuracies compared to scratch training with all of the formula-driven image datasets. The improvements are up to \cNaka{\{+18.5, +23.9, +74.4, +21.2\}} higher accuracies with FractalDB-1k on \{C10, C100, Cars, Flowers\} datasets. Note that the configuration is based on the original and standard FractalDB-1k which contains 1,000 [category] $\times$ 1,000 [instance]. In formula-driven image datasets, the FractalDB pre-trained \cHiro{DeiT} outperforms the other pre-trained models. \cHiro{The accuracies are \{+0.1, +1.3, +3.2, -0.2\} from BezierCurveDB} pre-trained \cHiro{DeiT}.
%All of the formula-based datasets show higher accuracy on fine-tuned datasets compared to learning from scratch, i.e. up to \{+18.3, +20.2, +74.6, +21.4\} higher accuracy from scratch on \{C10, C100, Cars, Flowers\}. 
%Among them, FractalDB outperforms the others as well as \cite{KataokaACCV2020}, i.e. \{+1.9, +3.7, +23.9, +2.4\} increase on \{C10, C100, Cars, Flowers\} from Perlin-324.
According to this result, we conduct the following experiments by FractalDB.

\noindent {{\bf\#category and \#instance (see Figures~\ref{fig:catins_c10}, \ref{fig:catins_c100}, \ref{fig:catins_cars}, \ref{fig:catins_flowers}).}  Figure~\ref{fig:catins} indicates the effects of increase for category and instance on FractalDB pre-training}. We set category and instance as variables, fixing one of them at 1000 and changing the others to \{16, 32, 64, 128, 256, 512, 1000\}. From the experimental results, a larger category and instance tend to lead to higher accuracy on a fine-tuning dataset. Especially in FractalDB pre-training, the category increase is more effective for transfer learning on an image dataset. 
%The accuracy is higher when category is changed than when instance is changed. 
This result is intuitive because the task is easier for datasets with fewer categories and more instances. 

Hereafter, we use 1,000 [category] $\times$ 1,000 [instance] as a basic setting of FractalDB. Due to the effectiveness of increasing the number of categories for improving the accuracy, we try to optimize 10,000 [category] $\times$ 1,000 [instance] as well. 
%Moreover, we additionally implement FractalDB-\{30k, 50k, 100k\} to further improve the transfer learning. In the case of FractalDB-\{30k, 50k, 100k\}, we set 500 instances per category due to the experimental results in Figure~\ref{fig:catins} and computational resources.
It is said to be better when the transformer's pre-training dataset is larger. \cHiro{We'd} like to confirm whether the tendency is applicable in image classification tasks.
%From then now, we assign 1,000 [category] $\times$ 1,000 [instance] as a basic dataset, and try to train on more than 10k categories as well, since a larger number of categories would contribute to better performance.

% DeiT-T/16 patch vs point
\begin{table}[t]
    \begin{center}
    \caption{Patch vs. point.}
    \begin{tabular}{lcccc} \toprule[0.8pt]
         & C10 & C100 & Cars & Flowers \\ \midrule[0.5pt]
        Point & 94.2 & 77.3 & 65.4 & 95.1 \\
        Patch & \textbf{96.8} & \textbf{81.6} & \textbf{86.0} & \textbf{98.3} \\ \bottomrule[0.8pt]
    \end{tabular}
    \label{tab:patchvspoint}
    \end{center}
    \vspace{-13pt}
\end{table}

% DeiT-T/16 grayscale vs. color
\begin{table}[t]
    \begin{center}
    \caption{Grayscale vs. color.}
    \begin{tabular}{lcccc} \toprule[0.8pt]
         & C10 & C100 & Cars & Flowers \\ \midrule[0.5pt]
        Grayscale & \textbf{97.1} & \textbf{82.6} & \textbf{87.1} & \textbf{98.3} \\
        Color & 96.8 & 81.6 & 86.0 & \textbf{98.3} \\ \bottomrule[0.8pt]
    \end{tabular}
    \label{tab:grayscale_vs_color}
    \end{center}
    \vspace{-13pt}
\end{table}

% DeiT-T/16 Training epochs
\begin{table}[t]
    \begin{center}
    \caption{\cNaka{Training epoch.}}
    \begin{tabular}{lcccc} \toprule[0.8pt]
        \#Epoch & C10 & C100 & Cars & Flowers \\ \midrule[0.5pt]
        100 & 96.1 & 81.1 & 82.0 & 96.5 \\
        200 & \textbf{96.8} & \textbf{82.1} & 85.3 & 98.2 \\
        300 & \textbf{96.8} & 81.6 & \textbf{86.0} & \textbf{98.3} \\ \bottomrule[0.8pt]
    \end{tabular}
    \label{tab:training_epochs}
    \end{center}
    \vspace{-20pt}
\end{table}

% DeiT-T パッチサイズ調査
\begin{table}[t]
    \begin{center}
    \caption{Patch size.}
    \begin{tabular}{lccccc} \toprule[0.8pt]
         & Size & C10 & C100 & Cars & Flowers \\ \midrule[0.5pt]
        \multirow{2}{*}{ImageNet-1k} & 16 & \textbf{98.0} & \textbf{85.5} & \textbf{89.9} & \textbf{99.4} \\
                     & 32 & 97.5 & 84.7 & 86.4 & 98.0 \\ \midrule[0.5pt]
        \multirow{2}{*}{FractalDB-1k} & 16 & \textbf{96.8} & \textbf{81.6} & \textbf{86.0} & \textbf{98.3} \\
                     & 32 & 95.5 & 78.4 & 76.0 & 95.7 \\ \bottomrule[0.8pt]
    \end{tabular}
    \label{tab:patchsize}
    \end{center}
    \vspace{-20pt}
\end{table}

\begin{table*}[t]
    \begin{center}
    \caption{\cNaka{Comparison of pre-training DeiT-Ti on several datasets. The optimization setting is based on \cite{TouvronarXiv2020}.}
    %Comparisons among paired pre-training datasets and models with FractalDB pre-trained ViT (ours), ImageNet-1k/Places-365 pre-trained ViT. The optimization method is based on DeiT-Ti.
    %Ours(FractalDB-1k/10k), Scratch, ImageNet-1k, Places-365 pre-trained models on representative pre-training datasets. The architecture is DeiT-T. 
    \cNaka{We show types of pre-trained image (PT img), which includes \{Natural Image (Natural), Formula-driven image dataset (Formula)\}; and pre-training type (PT Type), which includes \{Supervised learning (Supervision), Formula-driven supervised learning (Formula-supervision)\}.} We employed CIFAR-10 (C10), CIFAR-100 (C100), Stanford Cars (Cars), and Flowers-102 (Flowers), Pascal VOC 2012 (VOC12), Places-30 (P30), ImageNet-100 (IN100) datasets. The \textbf{\underline{Underlined bold}} and \textbf{bold} scores show the best and second best values, respectively.}
    \begin{tabular}{lccccccccc} \toprule[0.8pt]
        PT   & PT Img & PT Type & C10 & C100 & Cars & Flowers & VOC12 & P30 & IN100 \\ \midrule[0.5pt]
        %Scratch & ResNet & -- & 87.6 & 62.7 & **.* & **.* & 58.9 & 49.9 & 1.1 \\
        %Places-365 & ResNet & Natural & 94.2 & 76.9 & **.* & **.* & 78.6 & -- & 10.5 \\
        %ImageNet-1k & ResNet & Natural & 96.8 & 84.6 & **.* & **.* & 85.8 & 50.3 & 17.5 \\
        Scratch   & -- & -- & 78.3 & 57.7 & 11.6 & 77.1 & 64.8 & 75.7 & 73.2\\
        Places-30   & Natural & Supervision & \cTBD{95.2} & \cTBD{78.5} & \cTBD{69.4} & \cTBD{96.7} & \cTBD{77.6} & \cNaka{--} & \cTBD{86.5} \\
        Places-365   & Natural & Supervision & \textbf{97.6} & \textbf{83.9} & \textbf{89.2} & \textbf{99.3} & 84.6 & \cNaka{--} & \underline{\textbf{89.4}} \\
        ImageNet-100   & Natural & Supervision & \cTBD{94.7} & \cTBD{77.8} & \cTBD{67.4} & \cTBD{97.2} & \cTBD{78.8} & \cTBD{78.1} & \cNaka{--} \\
        ImageNet-1k   & Natural & Supervision & \underline{\textbf{98.0}} & \underline{\textbf{85.5}} & \underline{\textbf{89.9}} & \underline{\textbf{99.4}} & \underline{\textbf{88.7}} & \underline{\textbf{80.0}} & \cNaka{--} \\
        %FractalDB-30k~\cite{KataokaACCV2020} & ResNet & Formula & 91.1 & 76.2 & **.* & **.* & 56.1 & **.* & **.* \\ \hline
        \rowcolor[gray]{0.8}%
        FractalDB-1k   & Formula & Formula-supervision & 96.8 & 81.6 & 86.0 & 98.3 & 84.5 & 78.0 & 87.3 \\
        \rowcolor[gray]{0.8}%
        FractalDB-10k   & Formula & Formula-supervision & \textbf{97.6} & 83.5 & 87.7 & 98.8 & \textbf{86.9} & \textbf{\cTBD{78.5}} & \textbf{88.1} \\
        %\rowcolor[gray]{0.8}%
        %FractalDB-100k & DeiT & Formula & **.* & **.* & **.* & **.* & **.* & **.* & **.* \\
        \bottomrule[0.8pt]
    \end{tabular}
    \label{tab:comparison}
    \end{center}
    \vspace{-17pt}
\end{table*}

\noindent {{\bf Larger categories (see Table~\ref{tab:larger_category}).}  We conduct an experiment in larger categories on FractalDB. 
%Though FractalDB-10k is following the original one, we make the dataset with 500 instances per category in FractalDB-\{30k, 50k, 100k\}. Even in this case, the number itself is larger, e.g., FractalDB-10k with 1k instances per category contains 10M images vs. FractalDB-30k with 500 instances per category includes 15M images. 
The table indicates a larger FractalDB pre-training enhances the transformer in image classification. As a matter of fact, the accuracies are improved as \cNaka{\{96.8, 97.6\}} by FractalDB-\{1k, 10k\} pre-training on CIFAR-10.}

\noindent {\bf Patch vs. point (see Table~\ref{tab:patchvspoint}).} Table~\ref{tab:patchvspoint} indicates a comparison between $3 \times 3$ [pixel] patch rendering and $1 \times 1$ [pixel] point rendering. We execute the experiment to find a better way of fractal rendering. Though the point rendering represents a detailed pattern in a fractal image, the patch rendering augments the instances inside of the category. We can confirm that patch rendering increases performance with \{+2.6, +4.7, +20.6, +3.2\} on \{C10, C100, Cars, Flowers\}. We assign the patch rendering in FractalDB through the experiments.

\noindent {\bf Grayscale vs. color (see Table~\ref{tab:grayscale_vs_color}).} \cHiro{The table shows pre-training on FractalDB \cYama{works better with grayscale images than colored images}. Especially, in C100 and Cars datasets, the improvement gaps are +1.0 pt and +1.1 from the pre-training with colored fractal images. We confirmed that the colored representation \cYama{is} not required in \cHiro{DeiT} architecture.}

\noindent {\bf Training epoch (see Table~\ref{tab:training_epochs}).} \cHiro{In FractalDB pre-training, a longer training epoch tends to \cYama{achieve} better performance rates, \cYama{similarly to} SSL methods. The accuracies in 300 epoch pre-training recorded the best scores in three out of four different datasets.}

\noindent {\bf Patch size in \cHiro{DeiT} (see Table~\ref{tab:patchsize}).}
To input an image to \cHiro{DeiT}, an image is divided with multiple patches. Though the patch size in \cHiro{DeiT} is verified, we also seek a suitable patch size by comparing between ImageNet-1k and FractalDB-1k pre-training. As shown in Table~\ref{tab:patchsize}, we calculated \cHiro{DeiT} with different patch sizes, \{16$\times$16, 32$\times$32\} [pixel], at each pre-training setting. From the table, \cHiro{the 16$\times$16 patch size in both ImageNet-1k and FractalDB-1k is \cYama{the} better configuration} in three out of four datasets. % →フラクタル性との関係性を示す追加実験あれば「16x16が8x8より良いのはフラクタルの再帰性を捉えるちょうど良いサイズがあるのではないか？」というような書き込みを加えます

%\subsection{Comparison to natural image datasets}
%We compare the FractalDB constructed in Section~\ref{subsec:exploration} with large-scale natural image datasets (ImageNet-1K and Places-365). Table~\ref{tab:comparison} shows that FractalDB can achieve competitive accuracy with natural image datasets.
\subsection{Comparisons}
\label{sec:comparison}
We compare the performance of FractalDB pre-trained \cHiro{DeiT} with 
%FractalDB pre-trained ResNet-50~\cite{KataokaACCV2020}, 
\cHiro{\{ImageNet-1k, ImageNet-100, Places-365, Places-30\} pre-trained \cHiro{DeiT}} on representative datasets \cHiro{in addition to training from scrach with additional fine-tuning datasets}. \cHiro{ImageNet-100 and Places-30 are randomly selected categories from ImageNet-1k and Places-365 presented in~\cite{KataokaACCV2020}. Moreover, we also evaluate SSL methods with \{Jigsaw, Rotation, MoCov2, SimCLRv2\} on ImageNet-1k.} Here, we show the effectiveness of the proposed method in compared properties, namely \cHiro{human supervision with natural images (Table~\ref{tab:comparison}) and self supervision with natural images (Table~\ref{tab:fdsl_vs_ssl})}. %ResNet-50 vs. DeiT, and Natural images vs. Formula-driven images.

\noindent {\bf \cHiro{FDSL} vs. Supervised Learning.}
We compare natural image datasets and the FractalDB in the pre-training phase. \cHiro{Table~\ref{tab:comparison} describes the detailed settings in pre-training (PT), architecture (Arch.), and pre-training images (PT img) and their performance in terms of accuracy.} 
%In pre-training with natural image datasets, we employ ImageNet-1k and Places-365. 
\cHiro{At the beginning, the FractalDB-1k/10k pre-trained \cHiro{DeiT}s outperformed the pre-trained models on 100k-order labeled datasets (ImageNet-100 and Places-30). Although the FractalDB-10k pre-trained \cHiro{DeiT} did not exceed the performance with million-order labeled datasets (ImageNet-1k and Places-365), the scores are similar to the ImageNet-1k pre-trained \cYama{model}. }
%Although the relatively small-sized FractalDB-1k exhibits slightly lower performance rates, FractalDB-10k pre-trained ViT are comparable with the ImageNet-1k pre-trained one. 

\noindent {\bf \cHiro{FDSL vs. SSL.}} Through the comparisons with SimCLRv2, we clarify that the FractalDB-10k pre-trained \cHiro{DeiT} performs \cHiro{slightly higher (FractalDB-10k 88.8 vs. SimCLRv2 88.5) in \cYama{average} accuracy} on representative datasets.
%Through the comparisons with SimCLRv2, we clarify that the FractalDB-10k pre-trained ViT performs  comparatively on representative datasets.
The FractalDB-10k pre-training outperformed the SimCLRv2 pre-training on C10 (97.6 vs. 97.4), Cars (87.7 vs. 84.9), and VOC12 (86.9 vs. 86.2); the accuracy was lower on C100 (83.5 vs. 84.1), Flowers (98.8 vs. 98.9), and P30 (78.5 vs. 80.0).
\cHiro{In addition to SimCLRv2, we implemented Jigsaw, Rotation, MoCov2 to compare with our FractalDB-10k. Although the FractalDB-10k pre-trained model performs similarly to SimCLRv2, the proposed method recorded higher accuracies than other SSL methods including MoCov2, Rotation, and Jigsaw, except for P30 dataset.}
%\cHiro{By comparing to the other SSL methods, the proposed method surpassed the performance rates, except for P30 dataset.}
Further comparisons with other SSL methods are shown in Table~\ref{tab:fdsl_vs_ssl}.

\begin{comment}
\begin{figure*}[t]
    \centering
    \subfigure{\includegraphics[width=0.30\linewidth]{fig/filter/filters.png} \label{fig:filters}}
    \subfigure{\includegraphics[width=0.287\linewidth]{fig/pos_sim/deitt16_pos_sim.pdf} \label{fig:pos_sims}}
    \subfigure{\includegraphics[width=0.35\linewidth]{fig/mean_attention_distance/mean_attention_distance.pdf} \label{fig:att_dists}} \\
    \caption{\cHiro{\textbf{Left:} The filters of the first linear embedding on ImageNet-1k (top) and FractalDB-1k (bottom). \textbf{Center:} The similarity of positional embedding on ImageNet-1k (top) and FractalDB-1k (bottom).} \textbf{Right:} Mean attention distance between ImageNet-1k (top) and FractalDB-1k (bottom) pre-training.}
    \label{fig:viss}
\end{figure*}
\end{comment}

% \begin{figure}[t]
%     \centering
%     \subfigure{\includegraphics[width=0.3\linewidth]{fig/filter/filters.png} \label{fig:filters}}
%     \subfigure{\includegraphics[width=0.29\linewidth]{fig/pos_sim/deitt16_pos_sim.pdf} \label{fig:pos_sims}}
%     \subfigure{\includegraphics[width=0.355\linewidth]{fig/mean_attention_distance/mean_attention_distance.pdf} \label{fig:att_dists}} \\
%     \caption{\cHiro{\textbf{Left:} The filters of the first linear embedding on ImageNet-1k (top) and FractalDB-1k (bottom). \textbf{Center:} The similarity of positional embedding on ImageNet-1k (top) and FractalDB-1k (bottom).} \textbf{Right:} Mean attention distance between ImageNet-1k (top) and FractalDB-1k (bottom) pre-training.}
%     \label{fig:viss}
% \end{figure}

% FDSL vs. SSL
\begin{table*}[t]
    \begin{center}
    \caption{Detailed results with FDSL (FractalDB-10k) vs. SSL (Jigsaw, Rotation, MoCov2, SimCLRv2). Addition to columns also in Table 6, \cHiro{`Use Natural Images?' shows whether the natural images \cYama{was} used or not in the pre-training phase. `Average' indicates the \cYama{average} accuracy \cYama{of} all datasets in the table. ImageNet-100 is eliminated from the table because the listed SSL methods are trained by images on ImageNet-1k.} The \textbf{\underline{Underlined bold}} and \textbf{bold} scores show the best and second best values, respectively.}
    \begin{tabular}{lccccccc|c} \toprule[0.8pt]
        Method & \cHiro{Use Natural Images?} & C10 & C100 & Cars & Flowers & VOC12 & P30 & Average \\ \midrule[0.5pt]
        Jigsaw & \cHiro{YES} & 96.4 & 82.3 & 55.7 & 98.2 & 82.1 & \textbf{80.6} & 82.5 \\
        Rotation & \cHiro{YES} & 95.8 & 81.2 & 70.0 & 96.8 & 81.1 & 79.8 & 84.1 \\
        MoCov2 & \cHiro{YES} & 96.9 & 83.2 & 78.0 & 98.5 & 85.3 & \underline{\textbf{80.8}} & 87.1 \\
        SimCLRv2 & \cHiro{YES} & \textbf{97.4} & \underline{\textbf{84.1}} & \textbf{84.9} & \underline{\textbf{98.9}} & \textbf{86.2} & 80.0 & \textbf{88.5} \\
        %\rowcolor[gray]{0.8}
        %FractalDB-1k & Formula & 0 & 96.6 & 81.5 & \textbf{86.2} & 98.5 & 84.5 & 78.0 & 87.3 \\
        \rowcolor[gray]{0.8}
        FractalDB-10k & \cHiro{NO} & \underline{\textbf{97.6}} & \textbf{83.5} & \underline{\textbf{87.7}} & \textbf{98.8} & \underline{\textbf{86.9}} & 78.5 & \underline{\textbf{88.8}} \\
        \bottomrule[0.8pt]
    \end{tabular}
    \label{tab:fdsl_vs_ssl}
    \end{center}
    \vspace{-20pt}
\end{table*}

% \cHiro{DeiT} vs CNN (DeiT vs ResNet)
\begin{table}[t]
    \begin{center}
    \caption{DeiT vs. \cMatsu{ResNet} with FractalDB-1k pre-training.}
    \begin{tabular}{lccccc} \toprule[0.8pt]
        \multirow{2}{*}{Arch.} & Params & \multirow{2}{*}{C10} & \multirow{2}{*}{C100} & \multirow{2}{*}{Cars} & \multirow{2}{*}{Flowers} \\ 
        & \cHiro{(M)} & & & & \\ \midrule[0.5pt]
        ResNet-18 & 11 & 94.8 & 77.6 & 65.2 & 96.3 \\
        ResNet-34 & 21 & 95.9 & 79.4 & 79.8 & 84.9 \\
        ResNet-50 & 25 & 96.1 & 80.0 & 82.5 & 98.2 \\
        DeiT-Ti/16 & 5 & 96.8 & 81.6 & 86.0 & 98.3 \\
        DeiT-B/16 & 86 & 97.1 & 83.2 & 86.5 & 97.9 \\ \bottomrule[0.8pt]
    \end{tabular}
    \label{tab:vitvscnn}
    \end{center}
    \vspace{-20pt}
\end{table}

\subsection{Additional experiments}
\label{sec:additional_experiments}

\begin{figure}[t]
    \centering
    \subfigure[ImageNet-1k]{\includegraphics[width=0.29\linewidth]{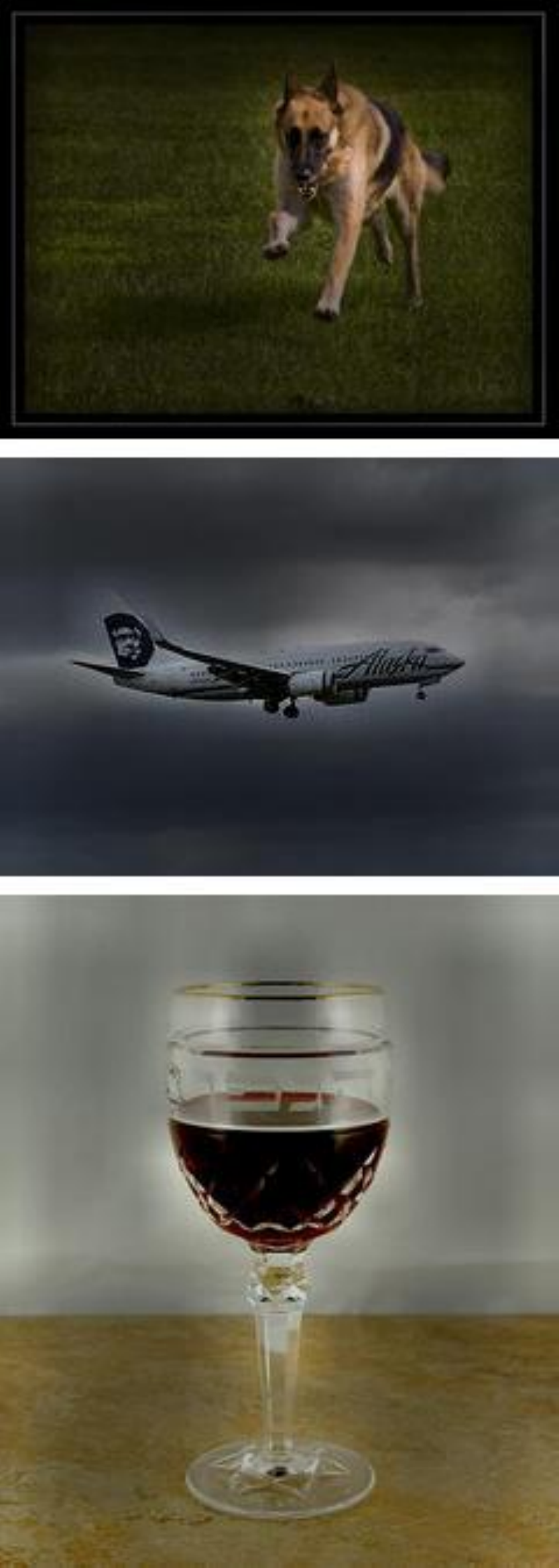} \label{fig:att_imagenet}}
    \subfigure[FractalDB-1k]{\includegraphics[width=0.29\linewidth]{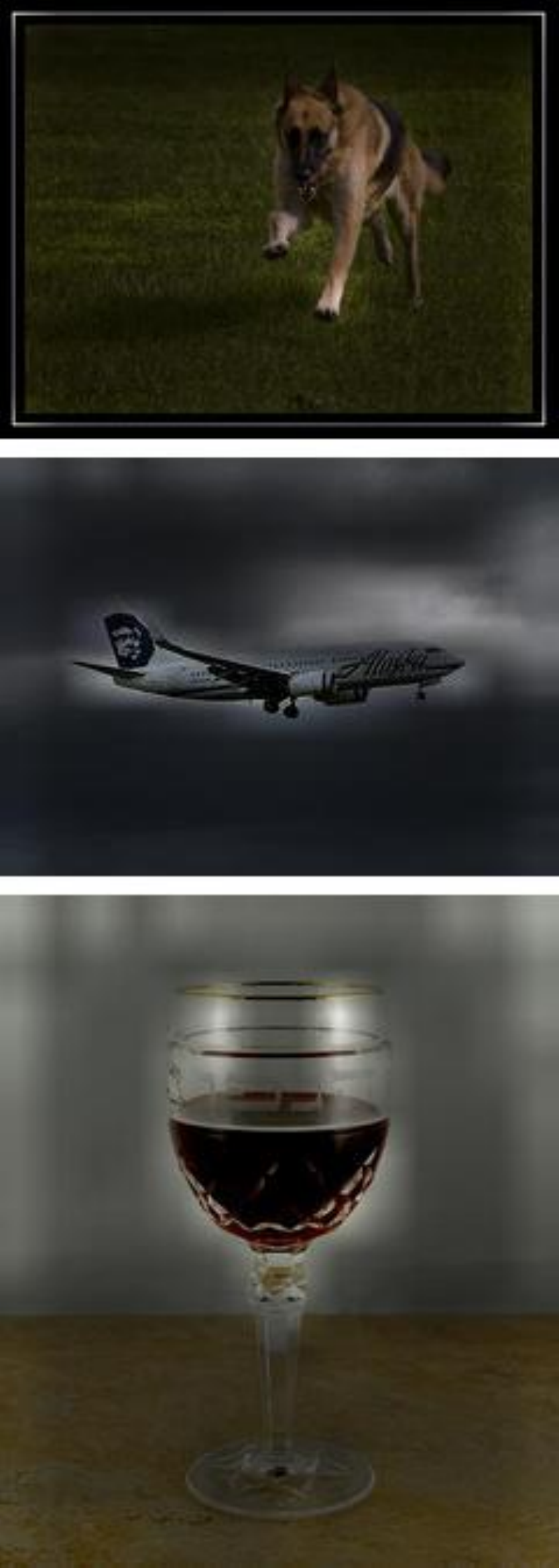} \label{fig:att_fractaldb1k}}
    \subfigure[FractalDB-10k]{\includegraphics[width=0.29\linewidth]{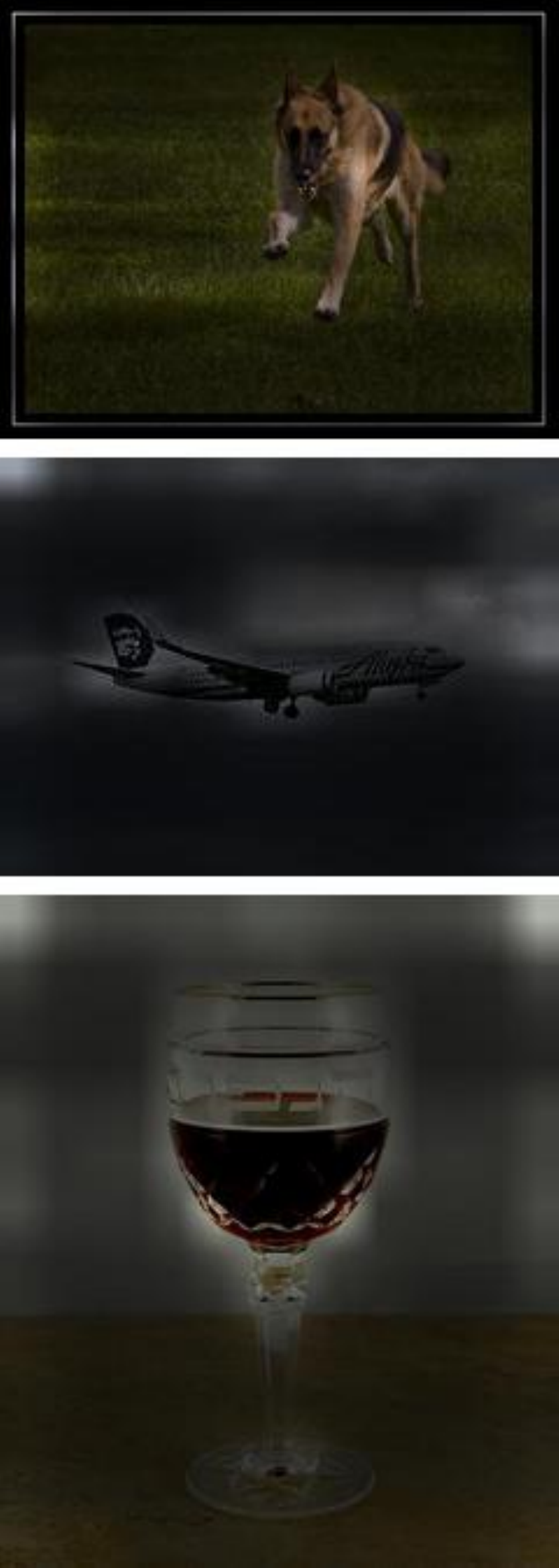} \label{fig:att_fractaldb10k}}
    \subfigure[Attention maps in fractal images with FractalDB-1k pre-trained \cHiro{DeiT}. \cHiro{The brighter areas show more attentive areas.} ]{\includegraphics[width=0.92\linewidth]{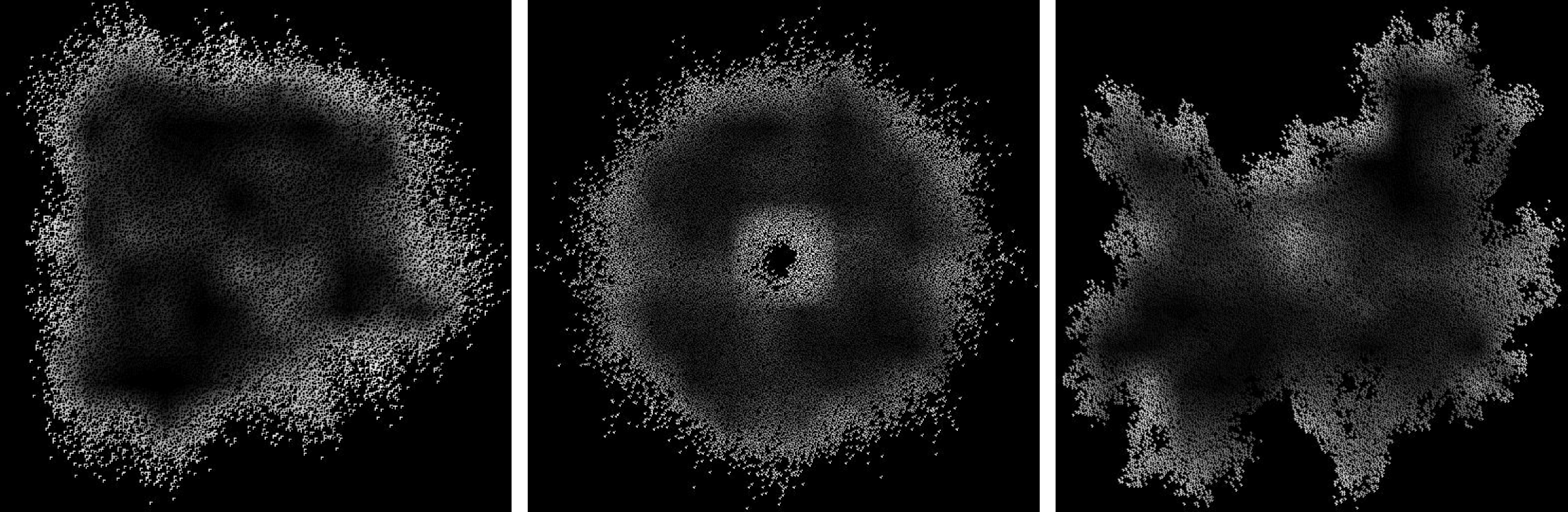} \label{fig:att_fractaldb1k_fractalimg}}
    \caption{Attention maps.}
    \label{fig:att}
    \vspace{-16pt}
\end{figure}

We conduct additional experiments in DeiT vs. ResNet (see Table~\ref{tab:vitvscnn}) by using more parameters. We also show the visualization through \cHiro{first linear embeddings, positional embeddings, mean attention distance \cMatsu{(Figure~\ref{fig:overview})}, and attention map (Figure~\ref{fig:att}).}

%\noindent {\bf FDSL vs. SSL (see Table~\ref{tab:fdsl_vs_ssl}).} In addition to SimCLRv2 in Table~\ref{tab:comparison}, we implemented Jigsaw, Rotation, MoCov2 to compare with our FractalDB-1k/10k. Although the FractalDB-10k pre-trained model performs similarly to  SimCLRv2, the proposed method recorded higher accuracies than other SSL methods including MoCov2, Rotation, and Jigsaw, for all fine-tuning datasets.

\noindent {\bf DeiT vs. ResNet (see Table~\ref{tab:vitvscnn}).}
We additionally verify DeiT and ResNet with different architecture sizes. We tested ResNet-\{18, 34, 50\} and DeiT-\{Ti, B\} with 16$\times$16 patch.
We assigned data augmentation in conjunction with the DeiT's setting.
The performance rates of ResNets and DeiTs are listed in Table~\ref{tab:vitvscnn}. 
At the beginning, different from \cHiro{the paper~\cite{KataokaACCV2020}}, the accuracies of ResNet-50 are better than previous ones (e.g., from 94.1 to 96.1 on C10).
%At the beginning, different from Table~\ref{tab:comparison} in Section~\ref{sec:comparison}, the accuracies of ResNet-50 are better than previous ones (e.g., from 94.1 to 96.1 on C10). 
However, the FractalDB pre-trained DeiTs are still better than FractalDB pre-trained ResNets on fine-tuning datasets.

%\noindent {\bf DeiT vs. ResNet.}
%We pre-trained FractalDB with the ResNet-50 and DeiT-Ti architectures. We summarize that the FractalDB pre-trained DeiT-Ti is better than FractalDB pre-trained ResNet-50 over all settings. 
%On one hand, the accuracy of ResNet-50 stopped increasing at FractalDB-10k. The FractalDB-30k pre-trained ResNet-50 does not exhibit the best accuracy in ResNet architecture. 

\noindent {{\bf Visualization (see Figure~\ref{fig:overview} and \ref{fig:att}).} 
For \cHiro{DeiT}, \cHiro{the filters of the first linear embedding, similarity of positional embedding, and mean attention distance can be visualized} by following the previous work~\cite{DosovitskiyICLR2021}.}
We list the filters as representations of ImageNet-1k and FractalDB-1k pre-trained models. %In \cHiro{DeiT}, divided image patches are mapped into a low-dimensional space. 
\cNaka{Figure~\ref{fig:overview}\cTab{(a)} shows} trained filters with ImageNet-1k and FractalDB-1k. Though both \cHiro{DeiT} pre-trained on ImageNet-1K and FractalDB-1K acquire similar filters, the FractalDB-1k pre-trained \cHiro{DeiT} tends to spread in wide-ranged areas of these filters. On one hand, the filters of ImageNet-1k pre-trained \cHiro{DeiT} seem to concentrate on center areas. %FractalDB is patch-rendered, but because it is sparser than natural images, the filter is generally blurred.
\cNaka{Figure~\ref{fig:overview}\cTab{(b)}} illustrates the cosine similarity of positional embedding corresponding to the input patch at each row and column. From the visualized figures, the FractalDB-1k pre-trained \cHiro{DeiT} acquired similar position embeddings at each row and column to ImageNet-1k pre-trained \cHiro{DeiT}. These pre-training datasets, ImageNet-1k and FractalDB-1k allow us to grab a feature from the same image position.
\cNaka{Figure~\ref{fig:overview}\cTab{(c)}} shows mean attention distance as in the original \cHiro{DeiT}~\cite{DosovitskiyICLR2021}. By comparing to the ImageNet-1k pre-training, the FractalDB-1k pre-trained \cHiro{DeiT} tends to look at wide-spread areas in an image. The indicator is similar to the size of receptive fields in CNN.

%Attention map
%\noindent {\bf Attention maps (see Figure~\ref{fig:att}).} 
\cHiro{Figure~\ref{fig:att} illustrates} attention maps in \cHiro{DeiT} with different pre-training datasets. The FractalDB-1k pre-trained ViT focuses on the object areas (Figure~\ref{fig:att_fractaldb1k}) as well as ImageNet pre-training (Figure~\ref{fig:att_imagenet}). Moreover, the FractalDB-10k pre-trained \cHiro{DeiT} looks at more specific areas \cNaka{(Figure~\ref{fig:att_fractaldb10k})} \cYama{compared} to FractalDB-1k pre-training. Figure~\ref{fig:att_fractaldb1k_fractalimg} shows attention maps in fractal images. From the figures, the FractalDB pre-training seems to \cYama{recognized} by observing contour lines. \cNaka{In relation to Figure~\ref{fig:overview}\cTab{(c)}}, we believe that the recognition of complex and distant contour lines enabled the extraction of features from a wide area.

\section{Conclusion \cHiro{and discussion}}
We successfully trained Vision Transformers (ViT) without any natural images \cHiro{and} human-annotated labels through the framework of Formula-Driven Supervised Learning (FDSL). Our FractalDB pre-trained ViT achieved similar performance rates to the human-annotated ImageNet pre-trained model, partially outperformed SimCLRv2 self-supervised ImageNet pre-trained model, and surpassed other self-supervised pre-training methods, including MoCov2. \cHiro{According to the results of the experiments, the findings are as follows.}

\noindent \cHiro{\textbf{Feature representation with FractalDB pre-trained ViT.} From the visualization results, the FractalDB pre-trained ViT \cYama{acquired different the feature representations in the first linear embeddings (Figure~\ref{fig:overview}\cTab{(a)}), and similar arranged position embeddings (Figure~\ref{fig:overview}\cTab{(b)}) compared to the ImageNet-1k pretrained model}.}
Moreover, \cHiro{Figure~\ref{fig:att_fractaldb1k_fractalimg} illustrates that ViT tends to pay attention to the contour areas in pre-training phase.} \cHiro{We believe that the pre-trained} model enabled feature acquisition in an area covering a wider range than the ImageNet-1k pre-trained model (Figure~\ref{fig:overview}\cTab{(c)}). We also understood the complex contour lines used to classify fractal categories in the pre-training phase. 

\noindent \cHiro{\textbf{Can we complete pre-training of ViT without natural images and human-annotated labels?}}
\cHiro{According to the comparisons with SSL methods (Table~\ref{tab:fdsl_vs_ssl}), we showed that the performance of FractalDB-10k was comparative to the accuracy of \cYama{SimCLRv2 pre-trained ViT, which is trained by 1.28M natural images on ImageNet}. Although 10M images are used in FractalDB-10k, natural images are not used at all in the pre-training phase. Therefore, we \cYama{can use a FDSL-based pre-training dataset to safely train ViT} in terms of AI ethics and image copyright, if we can exceed the accuracy of supervised learning with human annotation (Table~\ref{tab:comparison}).}

\section*{Acknowledgement}
\begin{itemize}
    \item This paper is based on results obtained from a project subsidized by the New Energy and Industrial Technology Development Organization (NEDO).
    \item Computational resource of AI Bridging Cloud Infrastructure (ABCI) provided by National Institute of Advanced Industrial Science and Technology (AIST) was used.
    \item We want to thank Yoshihiro Fukuhara, Munetaka Minoguchi, Eisuke Yamagata, Ryota Suzuki, Yutaka Satoh, Yue Qiu, and Shintaro Yamamoto for their helpful comments during research discussions.
\end{itemize}

{\small
\bibliographystyle{ieee_fullname}
\bibliography{egbib}
}

\end{document}